\documentclass{article}
\usepackage{spconf,amsmath,graphicx,bm}
\usepackage{multirow}
\usepackage{multicol}
\usepackage{graphicx}
\usepackage{subfigure} 
\usepackage{booktabs}
\usepackage{color}
\usepackage{hyperref}
\usepackage{balance}

\newcommand\eg{\emph{e.g.}}

\title{Human Activity Recognition Using 3D Orthogonally-projected EfficientNet on Radar Time-Range-Doppler Signature}
\name{Zeyu Wang$^{\star}$, Chenglin Yao$^{\star}$, Jianfeng Ren$^{\star}$, Xudong Jiang$^{\dagger}$
\thanks{This work was supported in part by the National Natural Science Foundation of China under Grant 72071116, and in part by the Ningbo Municipal Bureau Science and Technology under Grants 2019B10026.}}
\address{${}^{\star}$School of Computer Science, University of Nottingham Ningbo China\\${}^{\dagger}$Electrical \& Electronic Engineering, Nanyang Technological University}
\begin{document}
%
\maketitle
\begin{abstract}
In radar activity recognition, 2D signal representations such as spectrogram, cepstrum and cadence velocity diagram are often utilized, while range information is often neglected. In this work, we propose to utilize the 3D time-range-Doppler (TRD) representation, and design a 3D Orthogonally-Projected EfficientNet (3D-OPEN) to effectively capture the discriminant information embedded in the 3D TRD cubes for accurate classification. The proposed model aggregates the discriminant information from three orthogonal planes projected from the 3D feature space. It alleviates the difficulty of 3D CNNs in exploiting sparse semantic abstractions directly from the high-dimensional 3D representation. The proposed method is evaluated on the Millimeter-Wave Radar Walking Dataset. It significantly and consistently outperforms the state-of-the-art methods for radar activity recognition.


\end{abstract}
\section{Introduction}
\label{sec:intro}

\begin{figure*} [!ht]
	\centering
	\includegraphics[width=0.98\textwidth]{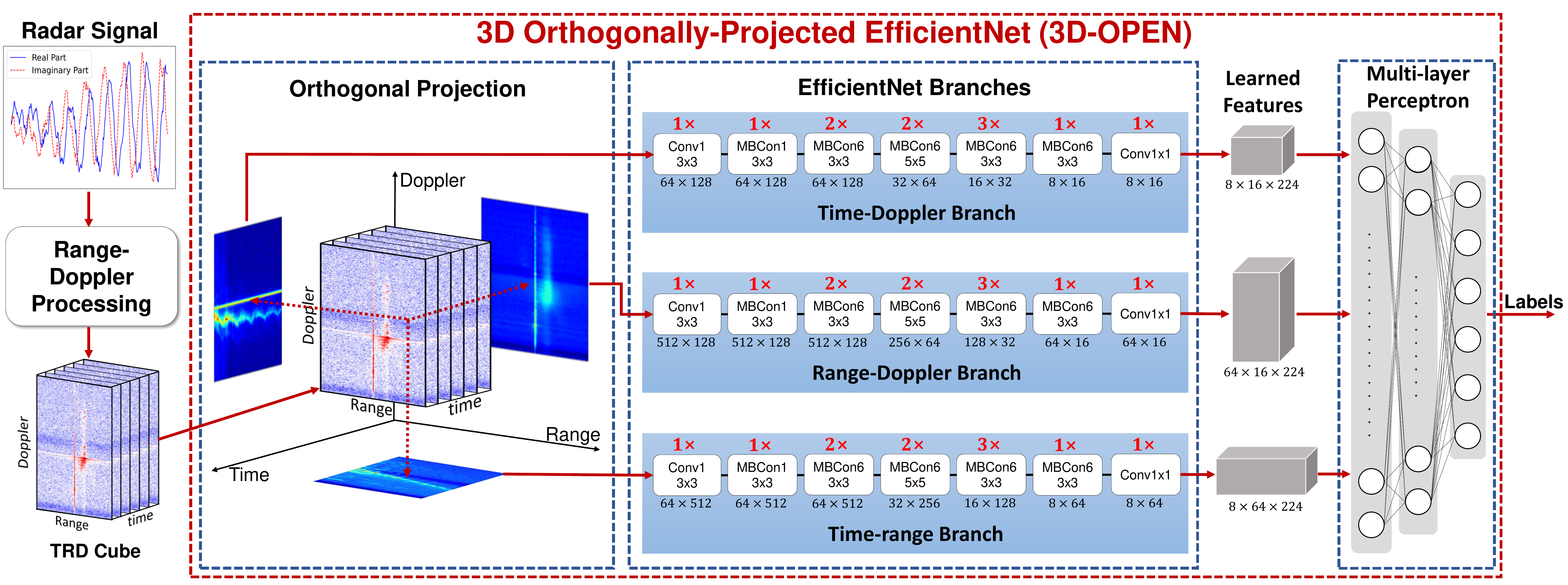}
	\caption{Overview of the proposed 3D-OPEN for human activity recognition using FMCW radars. The time-range-Doppler representation is first extracted from the radar signal and then classified using the proposed 3D-OPEN, in which the 3D TRD cubes are projected orthogonally onto three feature planes and processed using three customized EfficientNet branches. Three learned feature maps are reassembled in the Multi-layer Perceptron head for final prediction.}\label{fig:system}
	
\end{figure*}

Human activity recognition has advanced conspicuously with successful applications in surveillance, video retrieval and human-robot interaction over the past decade \cite{li2019survey}. Existing video-based methods for activity recognition have achieved promising results \cite{liu2014learning, iosifidis2014minimum, hara2018can}, but they are easily affected by lighting conditions, and capturing photos of users may raise privacy issues~\cite{li2019survey, meng2020gait}. Frequency-Modulated Continuous-Wave (FMCW) radar tracks the distance and velocity of targets~\cite{li2019survey} and has been successfully applied for remote sensing~\cite{ren2017regularized,ren2021three}. It could introduce robust and fine-grained sensing ability \cite{liu2020real}. In this paper, human activities are classified using FMCW radars, instead of using optical sensors.

FMCW radars capture the time, range and Doppler information of detected objects. The radar signal is often represented in 2D forms such as spectrogram~\cite{senigagliesi2020people, ahuja2021vid2doppler}, cepstrum \cite{wang2019rapid}, cadence velocity diagram~\cite{jia2020human} and time-range map~\cite{molchanov2015multi}, which may lose information either in range or Doppler frequencies. In literature, besides the one encoding the signal of a wearable radar sensor for hand gesture recognition~\cite{wang2016interacting}, very few methods utilize the 3D time-range-Doppler (TRD) representation. To encode the rich information in time, range and Doppler, we propose to utilize the 3D TRD presentation for human activity recognition.  

Many machine-learning models have been applied for human activity recognition, \eg, kNN~\cite{ding2019continuous}, SVM~\cite{senigagliesi2020people} and LDA \cite{erol2019radar}. The classification performance is often limited by the discriminant power of the models. CNN-based models such as AlexNet \cite{cao2018radar}, VGG-16~\cite{ahuja2021vid2doppler} and ResNet-18~\cite{addabbo2021temporal} have been applied on spectrograms for activity recognition, but the 2D signal representation limits the performance. Recently, 3D CNNs such as 3D ResNets have achieved competing classification results on video-based activity recognition \cite{hara2018can}. The spatio-temporal features could be effectively extracted from action videos using 3D convolutional networks.

However, the discriminant features are more sparsely spread in the 3D TRD representation than that in 3D videos, and that in the 2D radar signal representations, so that 3D CNNs could not effectively exploit high-level semantic abstractions from the 3D TRD representation. The computation load of 3D CNNs is also significantly heavier than that of 2D CNNs. To cope with these challenges, this paper proposes a 3D Orthogonally-Projected EfficientNet (3D-OPEN). Taking the 3D TRD as the input, the model firstly projects the data cube orthogonally to three feature planes. Then, the three compressed representations are fed into three network branches built using customized EfficientNet~\cite{tan2019efficientnet}. Benefited from its compound-scaling techniques, the three network branches can efficiently extract the discriminant information from the 3D TRD representation. The learned features are fed into a Multi-Layer Perceptron for final classification.

The proposed method is compared with the state-of-the-art methods~\cite{ahuja2021vid2doppler, wang2016interacting, cao2018radar, addabbo2021temporal, wang2018non} on the Millimeter-Wave Radar Walking Dataset (mmWRWD)~\cite{senigagliesi2020people,gambi2020millimeter}. It significantly and consistently outperforms all compared approaches. 

\section{Proposed Method}

\subsection{Overview of Proposed Method}
The proposed system for human activity recognition using FMCW radars is shown in Fig.~\ref{fig:system}. The radar signal is first transformed to time-range-Doppler cubes through range-Doppler processing~\cite{ding2019continuous}. Then, the generated 3D cubes are fed into the proposed 3D Orthogonally-Projected EfficientNet for activity recognition. The model firstly projects the cubes along the time, range and Doppler axes into range-Doppler, time-Doppler and time-range planes, respectively. Then, the projected representations in three planes are inputted separately into three EfficientNet branches to extract the discriminant features. The network architectures of the three branches are shown in Fig~\ref{fig:system}. Finally, the learned features are fed through the Multi-layer Perceptron for classification.

\subsection{Time-Range-Doppler Representation}
Many signal representations have been utilized for radar activity recognition \cite{li2019survey}, \eg, 3D time-range-Doppler cubes \cite{wang2016interacting} and 2D signal representations such as spectrogram~\cite{senigagliesi2020people, ahuja2021vid2doppler} and its variants cepstrogram~\cite{wang2019rapid} and cadence velocity diagram~\cite{jia2020human}. Spectrograms are more commonly used because they are intuitive and explicable~\cite{li2019survey}, which could be transformed from primitive radar signals by applying Short Time Fourier Transform (STFT) on each range bin and adding them up across the range bins of FMCW radars~\cite{li2019survey, senigagliesi2020people}. 

FMCW radars could well capture the time, range and Doppler information of the target. In this work, we propose to utilize the 3D time-range-Doppler representation to well encode the discriminant information of the radar signal for human activity recognition. Denote the radar signal as $\bm{s}$ and TRD representation as $\bm{R}_{TRD}$. $\bm{R}_{TRD}$ can be generated through range-Doppler processing $\mathcal{F}_{RDP}$~\cite{ding2019continuous} as:
\begin{equation}
\bm{R}_{TRD} = \mathcal{F_{RDP}}(\bm{s}).
\end{equation}
$\bm{R}_{TRD}$ could well capture the information of moving targets, \eg, utilizing the radar chirps to determine target's distance and the Doppler effect to determine target's velocity~\cite{li2019survey, liu2020real}.  However, time-range-Doppler cubes are rarely used in literature due to the complicated 3D structure and increased intra-class variations in the high-dimensional space.

Specifically, 
the FMCW radar signal is first cut into chirps of $M$ samples and stacked to form a data matrix of size $M\times NT$, where each row is a chirp and each column corresponds to a range bin. The data matrix is then divided into $T$ frames, where each contains $N$ successive chirps. Finally, the Fast Fourier transform (FFT) is applied on each chirp to derive the range information, and applied on each range bin of a frame to generate the Doppler signatures, in the same way as in~\cite{senigagliesi2020people}. 

The resulting $\bm{R}_{TRD} \in \mathcal{R}^{T\times M\times N}$ contain various information of the target, \eg, the speed estimated from the Doppler signatures, the distance away from the radar and the exact time instance. Each range-Doppler frame reveals the relative ranges and velocities of detected objects while the temporal information is embedded in the successive frames~\cite{li2019survey}. Compared to other representations, it fully reserves the radar features in time, range and Doppler, so that the subsequent CNN models could extract them for better recognition. As human activities contain different relative distances and speeds of body parts, the time-range-Doppler cubes characterize various patterns of human activities.

\subsection{Proposed 3D Orthogonally-Projected EfficientNet for Time-Range-Doppler Representation}

The 3D TRD representation contains more information than 2D representations such as spectrogram~\cite{senigagliesi2020people, ahuja2021vid2doppler}, CVD~\cite{jia2020human} and cepstrogram~\cite{wang2019rapid}, but its size is significantly increased. Some 3D CNNs such as 3D ResNet~\cite{wang2018non} and 3D GoogleEtE \cite{wang2016interacting} have been used to exploit the features in time, range, and Doppler dimensions simultaneously, but two challenges remain. Firstly, the additional dimension not only increases the intra-class variations, but also makes the features in the 3D TRD representation sparser than that in the 2D representations, so that the discriminant information in the 3D TRD representation are difficult to be extracted by directly applying the 3D CNNs~\cite{wang2016interacting, wang2018non}. Secondly, to handle the 3D representations, the computation load of the 3D CNNs~\cite{wang2016interacting, wang2018non} is enormously heavier than those 2D CNNs~\cite{ahuja2021vid2doppler,cao2018radar,addabbo2021temporal}. 

To effectively aggregate the discriminant features from the 3D TRD representation, a 3D Orthogonally-Projected EfficientNet is designed. As shown in Fig.~\ref{fig:system}, it first projects the 3D TRD cubes $\bm{R}_{TRD}$ along the time, range and Doppler axes into three orthogonal 2D planes using average pooling. Formally, denote the orthogonal projections along the time, range and Doppler axes as $\mathcal{P}_t$, $\mathcal{P}_r$ and $\mathcal{P}_D$, respectively. The features projected onto the range-Doppler, time-Doppler and time-range planes are respectively derived as,
\begin{align}
\bm{F}_{rD} = \mathcal{P}_t(\bm{R}_{TRD}) \in\mathcal{R}^{M\times N},\\
\bm{F}_{tD} = \mathcal{P}_r(\bm{R}_{TRD}) \in\mathcal{R}^{T\times N},\\
\bm{F}_{tr} = \mathcal{P}_D(\bm{R}_{TRD}) \in\mathcal{R}^{T\times M}.
\end{align}
The discriminant information embedded in $\bm{R}_{TRD}$ could be optimally projected into these three planes by using the orthogonal projections, and the extracted features $\bm{F}_{rD}$, $\bm{F}_{tD}$ and $\bm{F}_{tr}$ are physically interpretable. Large networks such as ResNet-50 and ResNet-101 may suffer from severe overfitting due to the limited number of training samples in radar applications. In view of the power of compound-scaling techniques of EfficientNets~\cite{tan2019efficientnet}, the three network branches, $\mathcal{B}_{rD}$ for range-Doppler plane, $\mathcal{B}_{tD}$ for time-Doppler plane and $\mathcal{B}_{tr}$ for time-range plane, are constructed using light-weight EfficientNet to learn the high-level human activity abstractions. The depth and width of the EfficientNet in each branch are adjustable to account for the variations in resolution. The detailed architectures of $\mathcal{B}_{rD}$, $\mathcal{B}_{tD}$ and $\mathcal{B}_{tr}$ are shown in Fig.~\ref{fig:system}, which are empirically optimized through Network Architecture Search to obtain the best classification performance. The output features are generated from these three branches as,
\begin{align}
\bm{O}_{rD} = \mathcal{B}_{rD}(\bm{F}_{rD}), \\
\bm{O}_{tD} = \mathcal{B}_{tD}(\bm{F}_{tD}), \\
\bm{O}_{tr} = \mathcal{B}_{tr}(\bm{F}_{tr}).
\end{align}
As these output features have different sizes, they are flatten and concatenated to form the fused feature $\bm{O}_{f}$. Finally, a Multi-Layer Perceptron $\mathcal{M}$ is utilized for final classification,
\begin{align}
\hat{l} = \mathcal{M}(\bm{O}_{f}),
\end{align}
where $\hat{l}$ is the predicted class label. The stochastic gradient descent optimizer is used for model training, and the cross entropy is utilized as the loss function.

The proposed 3D-OPEN could extract more discriminant features from the enriched 3D TRD representation than 2D networks on 2D representations. Compared to directly applying 3D networks such as 3D ResNet~\cite{wang2018non} and 3D GoogleEtE~\cite{wang2016interacting}, the proposed 3D-OPEN could well condense the discriminant information embedded in the 3D TRD representation into three orthogonal planes and effectively extract it using the customized EfficientNet branches. The compound-scaling mechanism of EfficientNet could also help to address the small-sample-size problem for radar activity recognition.

\section{Experimental Results}

\subsection{Experimental Settings}




The proposed model is compared with the state-of-the-art models on the Millimeter Wave Radar Walking Dataset~\cite{senigagliesi2020people, gambi2020millimeter}, including 2D networks such as AlexNet \cite{cao2018radar}, VGG-16 \cite{ahuja2021vid2doppler} and ResNet-18 \cite{addabbo2021temporal} on spectrogram, and 3D networks such as GoogleEtE \cite{wang2016interacting} and nonlocal ResNet-18 \cite{liao2018video} on the 3D TRD representation. The dataset consists of $231$ tests of $29$ people performing $6$ activities: fast walking (Fast), slow walking (Slow), slow walking with hands inside pockets (SlowPocket), walk with a limp (Limping), walk with a metallic bottle under the jacket (HidingBottle), and walk at slow speed with swinging hands (Swinging). Each test contains $4$ sequences concurrently collected from $4$ antennas. Each sample sequence consists of $400$ range-Doppler frames, $128$ chirps per frame, and $512$ sampling points per chirp, forming time-range-Doppler cubes of $400\times512\times128$ through range-Doppler processing. To enrich the dataset for robust training of CNNs, the $400$-frame cubes are evenly cropped into six $64$-frame cubes, which gives the input size of $64\times512\times128$ to the network. In total, $5544$ 3D TRD cubes are utilized in the experiments.

The Leave-One-Out-Cross-Validation protocol is used. In each fold, the samples collected from $3$ antennas are used as the training set while the samples from the remaining $1$ antenna are used as the test set. The reported classification accuracy are averaged over $4$ folds. Under this protocol, the model performance on unknown FMCW radars/antennas can be evaluated. The models are trained on two NVIDIA V100 GPUs with the initial learning rate at $0.1$ and an adaptive batch size. The proposed model converges after 300 epochs and reaches the optima within 600 epochs.
 

\subsection{Ablation Study}
To evaluate each contributing component of the proposed method, an ablation study is performed. As the features on each orthogonal plane are learned via the EfficientNet branch, the baseline method using EfficientNet-B0 on spectrogram is compared to show the performance of EfficientNet on 2D representations. An 3D EfficientNet, adapted from EfficientNet-B0 by expanding 2D convolution kernels and 2D pooling layers to 3D, is applied on the 3D TRD representation to assess the performance of 3D network on 3D TRD. The results are summarized in Table \ref{tab:results2d3d}.
\begin{table}[htbp!]
\centering 
\caption{Ablation study. The proposed 3D-OPEN on TRD significantly outperforms 3D EfficientNet on TRD and 2D EfficientNet on spectrogram. Note that directly applying 3D EfficientNet on the 3D TRD representation produces even poorer performance than 2D EfficientNet on spectrogram.}
\begin{tabular}{c|c|c}
\toprule
Dim. & Model & Accuracy \\
\midrule
2D & EfficientNet-B0 & $85.9\%$  \\
3D & EfficientNet & $77.7\%$ \\
3D & Proposed 3D-OPEN & $\bm{91.6\%}$ \\
\bottomrule
\end{tabular} \label{tab:results2d3d} 
\end{table}

The experimental results in Table \ref{tab:results2d3d} verify the superiority of the proposed 3D-OPEN on the 3D TRD representation. Compared to 2D EfficientNet on spectrogram, the performance gain is $5.7\%$, which shows that the 3D TRD representation contains more discriminant information and the proposed 3D-OPEN could effectively extract this informamtion. Directly applying 3D EfficientNet on TRD yields even poorer performance than 2D EfficientNet on spectrogram, which shows the difficulty in extracting the discriminant information from the 3D TRD representation. The proposed method significantly outperforms 3D EfficientNet by $13.9\%$, which demonstrates the effectiveness of the proposed 3D-OPEN in handling the 3D TRD representation.

\subsection{Comparison to State-of-the-art Methods}
The proposed method is compared to five state-of-the-art methods. All existing methods are reproduced and evaluated on the mmWRWD~\cite{senigagliesi2020people, gambi2020millimeter}. The hyper-parameters such as initial learning rate and the batch size are customized for the best performance for each method. The comparison results are summarized in Table \ref{tab:results-sota}. 
\begin{table}[htbp!]
	\centering 
	\caption{Comparison to the state-of-the-art models~\cite{ahuja2021vid2doppler, wang2016interacting, cao2018radar, addabbo2021temporal, wang2018non} for radar activity recognition. The proposed 3D-OPEN significantly outperforms all the compared methods.}
		
		\begin{tabular}{c|c|c}
			\toprule
			Dim. & Model & Accuracy \\
			\midrule
			2D & AlexNet \cite{cao2018radar}  & $41.2\%$  \\
			2D & VGG-16 \cite{ahuja2021vid2doppler} & $81.8\%$ \\
			2D & ResNet-18 \cite{addabbo2021temporal} &  $82.5\%$ \\
			\midrule
			3D & GoogleEtE \cite{wang2016interacting} & $83.2\%$ \\
			3D & ResNet-18+Non-local \cite{liao2018video} & $79.7\%$  \\
			\midrule
			3D & Proposed 3D-OPEN & $\bm{91.6\%}$  \\
			\bottomrule
		\end{tabular} \label{tab:results-sota} 
\end{table}

\begin{figure} [!ht]
	\centering
	\includegraphics[width=0.35\textwidth]{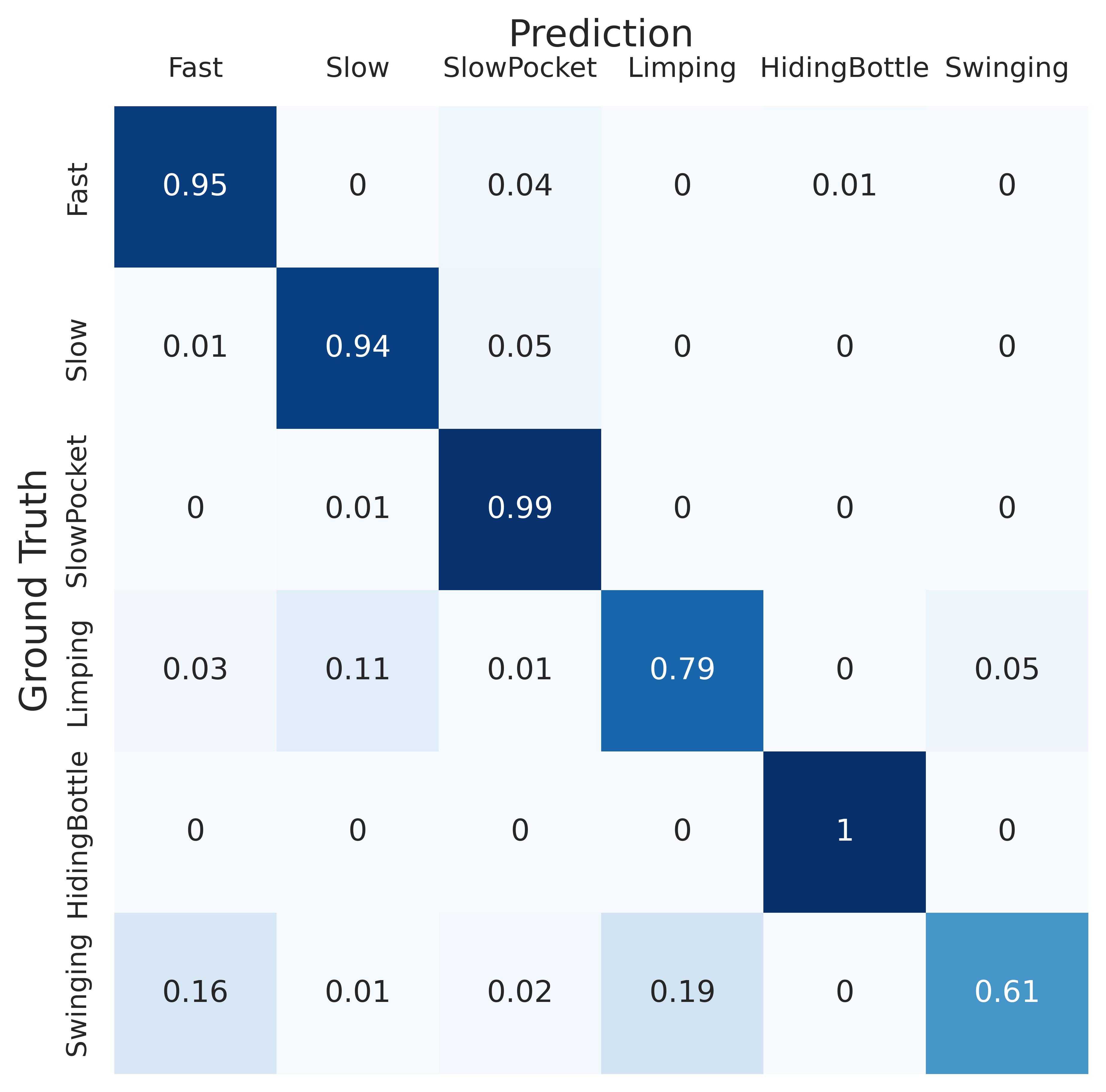}
	\caption{Confusion matrix of the proposed 3D-OPEN. ``Limping" and ``Swinging" may be wrongly interpreted as other activities, while the rest activities can be accurately classified.}
	\label{fig:cm}	
\end{figure}

The experimental results reveal that the proposed method significantly outperforms all the compared methods. The proposed method outperforms the best-performed existing method on 2D representation, 2D ResNet-18~\cite{addabbo2021temporal}, by $9.1\%$, and outperforms the best-performed existing method on 3D representation, GoogleEtE \cite{wang2016interacting}, by $8.4\%$. It is worth to note that 3D ResNet-18 with nonlocal blocks on the 3D TRD representations performs poorer than 2D ResNet-18 on spectrogram. All these demonstrate the superiority of the proposed method in extracting the discriminant information from the 3D TRD representation. The confusion matrix of the proposed method is shown in Fig.~\ref{fig:cm}. It can be seen that activity ``Limping" and ``Swinging" are easily misclassified, whereas others such as ``HidingBottle" can be accurately classified.  


\section{Conclusion}
In this work, we propose a 3D Orthogonally-Projected EfficientNet on 3D time-range-Doppler representation for human activity classification using FMCW radars. The proposed method addresses the problem of information loss of 2D radar signal representations and the challenges of feature learning from the 3D representation using 3D CNNs. The proposed method is evaluated on the Millimeter Wave Radar Walking Dataset. Compared to the state-of-the-art methods based on 2D CNNs and 3D CNNs, the proposed method achieves the significant and consistent performance gain.

\vfill\pagebreak

\newpage

\small
\balance
\bibliographystyle{IEEEbib}
\bibliography{refs,misc,jfR,additional}

\end{document}